\documentclass[11pt]{article}

\usepackage[margin=1in]{geometry}
\usepackage[utf8]{inputenc}
\usepackage[T1]{fontenc}
\usepackage{graphicx}
\usepackage{booktabs}
\usepackage{tabularx}
\usepackage{multirow}
\usepackage{amsmath}
\usepackage{amssymb}
\usepackage{microtype}
\usepackage{xcolor}
\usepackage{caption}
\usepackage{url}
\usepackage[hidelinks]{hyperref}

\title{\textbf{Deep Learning for Detecting Building Setback Rule Violations in Urban Planning: A Case Study on Bangladesh}}

\author{
Farzana Tasnim$^{1}$,\quad
Faria Yasmin$^{1}$,\quad
Md.\ Tauseef-Ur-Rahman$^{1}$, \\
Amina Islam$^{1}$,\quad
Tamzid Bin Tariq$^{1}$,\quad
Yuki Bhuiyan$^{1}$,\quad
Muhammad Ibrahim$^{1*}$ \\[3mm]
$^{1}$Department of Computer Science and Engineering, University of Dhaka, \\
Dhaka-1000, Bangladesh. \\[5mm]
\small
$^{*}$Corresponding author. E-mail:
\href{mailto:ibrahim313@du.ac.bd}
{\textcolor{blue}{ibrahim313@du.ac.bd}}; \\[1mm]
E-mail(s):
\href{mailto:farzanatasnim2003@gmail.com}
{\textcolor{blue}{farzanatasnim2003@gmail.com}};
\href{mailto:fariayasmin19@gmail.com}
{\textcolor{blue}{fariayasmin19@gmail.com}}; \\
\href{mailto:mdtauseef.rahmang01@gmail.com}
{\textcolor{blue}{mdtauseef.rahmang01@gmail.com}};
\href{mailto:aminaislam1912@gmail.com}
{\textcolor{blue}{aminaislam1912@gmail.com}}; \\
\href{mailto:tamzidbintariq@gmail.com}
{\textcolor{blue}{tamzidbintariq@gmail.com}};
\href{mailto:yukibhuiyan9291@gmail.com}
{\textcolor{blue}{yukibhuiyan9291@gmail.com}};
}

\date{}

\begin{document}

\maketitle
\begin{abstract}
Dhaka's rapid, largely unregulated urban growth has made building setback violations common, narrowing footpaths, blocking emergency access, and degrading the streetscape in neighborhoods across the city. RAJUK, the authority responsible for building control in the Dhaka Metropolitan Area, enforces setback rules mainly through manual field inspection, an approach that scales poorly across a city of this size. We present a systematic comparison of four machine learning approaches for classifying building setback conditions from street-level photographs into three categories: negative (compliant), moderate (borderline), and positive (violation). We built a labeled dataset of 372 geo-tagged images from sixteen Dhaka neighborhoods, each assigned a class label and a measured building-to-boundary distance, and expanded it with 5,968 offline augmented copies for training. On this dataset we trained and evaluated, under a leakage-free, subject-level train/validation/test split: a fully-connected multi-layer perceptron (MLP) operating on raw pixels as a lower-bound baseline; a heterogeneous soft-voting ensemble of Random Forest, Extra Trees, and logistic regression classifiers on frozen ResNet50 embeddings; a bagged linear support vector machine trained on class-balanced samples of frozen ResNet18 embeddings; and a ResNet18 convolutional network fine-tuned end to end with hard-example mining. On a shared 1,275-image held-out test set, the fine-tuned CNN reached the highest test accuracy, 65.4\% (macro-F1 0.50), followed by the Random Forest ensemble at 59.9\% (macro-F1 0.44) and the balanced SVM ensemble at 58.2\% (macro-F1 0.45); the raw-pixel MLP, evaluated on its own smaller 56-image real-only test set, reached 50.0\% (macro-F1 0.42). All four models classify the compliant class reliably but struggle with the borderline moderate class, which we attribute to its inherent visual ambiguity and its smaller share of the labeled data. These results indicate that end-to-end convolutional feature learning provides a measurable advantage over classical ensembles built on frozen embeddings, while both substantially outperform a pixel-based baseline, and they establish reference performance levels for automated setback-compliance screening in a South Asian regulatory setting.
\end{abstract}

\section{Introduction}

Dhaka is one of the fastest-growing megacities in the world, with a population that has passed 22 million and a built environment expanding faster than the city's planning systems can track \cite{worldbank2023}. One consequence is widespread disregard for construction regulations, and violations of building setback rules are among the most visible and damaging. RAJUK (Rajdhani Unnayan Kartripakkha), the authority responsible for guiding construction across greater Dhaka, has long sought practical ways to bring this problem under control \cite{rajuk2008}.

A building setback is the minimum legally required gap between a structure and its plot boundary or the adjacent road. Under RAJUK's Dhaka Metropolitan Building Construction Rules and the Bangladesh National Building Code, the required distance depends on road width, land-use zone, and building height \cite{bnbc2020}. When a building does not respect this distance, it encroaches on public footpaths and rights-of-way, with predictable consequences: reduced pedestrian safety, blocked emergency access, and a deteriorating streetscape.

Manual field inspection, RAJUK's primary enforcement tool today, is slow, resource-intensive, and inconsistent between inspectors. An automated system that can screen a large number of street-level photographs quickly and consistently would let field staff prioritize which buildings to inspect in person, rather than replacing inspection altogether.

Deep learning has already shown strong results in comparable settings, including agriculture \cite{ferdaus2025,tahsin2025}, road infrastructure \cite{hossen2025}, and urban analytics \cite{lubaina2025}, often relying on transfer learning when labeled data is limited \cite{hasan2024}. We are not aware of prior work that targets building setback violation detection from street-level imagery in a South Asian regulatory context specifically. Most existing studies on urban regulatory violations rely on aerial or satellite imagery \cite{huang2021}, which cannot capture the facade-to-road proximity that defines a setback violation at ground level.

This paper extends our earlier work on this problem in two ways. First, we add two classical machine learning pipelines, a heterogeneous Random Forest ensemble and a class-balanced support vector machine, both operating on frozen convolutional embeddings, to the deep learning models we studied previously. Second, we report train, validation, and test accuracy for every model under a consistent, leakage-free evaluation protocol, so the contribution of representation learning (raw pixels, frozen embeddings, or end-to-end fine-tuning) can be assessed directly. Our contributions are as follows.

\begin{itemize}
\item \textbf{A labeled, real-world dataset.} We conducted on-foot surveys across sixteen Dhaka neighborhoods, collecting and labeling 372 building facade images as negative (compliant), moderate (borderline), or positive (violation), each with a measured building-to-boundary setback distance.
\item \textbf{A four-way model comparison under one protocol.} We benchmark a from-scratch MLP, a Random Forest heterogeneous ensemble, a balanced bagged SVM, and a ResNet18 transfer-learning CNN, all evaluated with subject-level, leakage-free splitting so that near-duplicate augmented images never cross between training and evaluation sets.
\item \textbf{An analysis of representation strategy.} By holding the dataset and the general evaluation methodology fixed while varying only how each model represents an image (raw pixels, frozen embeddings, or fine-tuned features), we quantify how much of the classification performance comes from learned spatial representations rather than from the classifier on top of them.
\item \textbf{A practical deployment pathway.} The data collection method, smartphone photographs taken during ordinary field surveys, requires no specialized equipment and fits into RAJUK's existing inspection workflow, so the trained models could plausibly be packaged into a mobile or web tool for field staff.
\end{itemize}

\section{Related Work}

\textbf{Footpath and sidewalk encroachment detection.} The most closely related work is FootpathVision \cite{lubaina2025}, the first publicly available labeled dataset for sidewalk encroachment detection in Dhaka. It contains 1,238 images with COCO-style bounding box annotations supporting both binary (encroached versus clear) and multi-class (illegal stalls, illegal parking, clear) tasks. Evaluations across MobileNetV2, EfficientNetB0, ResNet50, VGG16, and DenseNet121 showed that binary classification is robust while multi-class detection is harder under class imbalance, a pattern we also observe for the moderate class in this work.

\textbf{Building and construction violation detection.} Most existing work on illegal construction relies on satellite or aerial imagery. Huang et al.\ \cite{huang2021} used multi-temporal satellite images and change detection to flag unauthorized construction, and Varghese et al.\ proposed ChangeNet \cite{varghese2018} for visual change detection in public urban spaces. These methods cannot capture facade-to-road proximity, which is why street-level imagery is necessary for setback assessment specifically.

\textbf{Road and infrastructure damage detection.} Hossen et al.\ \cite{hossen2025} built a YOLOv9 pipeline with polygonal annotations for road damage and manhole detection in Dhaka, reaching an 86.7\% F1-score for broken-road detection on a dataset of over 1,000 images, and Maeda et al.\ \cite{maeda2018} showed that smartphone-captured road images support effective deep-learning-based damage detection. Both results support the practicality of smartphone-based, locally collected data for Bangladeshi infrastructure monitoring, the same collection strategy we use here.

\textbf{Sidewalk accessibility.} The Project Sidewalk dataset \cite{weld2019}, with over 300,000 image-based labels, showed that modified ResNet architectures approach human-level performance in identifying accessibility problems from street-view panoramas. Hosseini et al.\ \cite{hosseini2022} combined crowd-sourced data with AI-assisted sidewalk mapping to extend this direction.

\textbf{Transfer learning and classical ensembles for small datasets.} With large-scale annotation costly and slow, fine-tuning ImageNet-pretrained backbones is the standard approach when labeled data is scarce. Residual connections \cite{he2016} make ResNet a natural choice for partial fine-tuning; MobileNetV2 \cite{sandler2018} and EfficientNet \cite{tan2019} offer lighter alternatives suited to constrained deployment, and Hasan et al.\ \cite{hasan2024} showed that genetic-algorithm-guided selection of which blocks to update can further improve transfer learning on domain-specific tasks. Separately, ensembles of classical learners built on top of frozen deep features, rather than end-to-end fine-tuning, remain a common and often competitive strategy when the labeled sample is small. Random Forests \cite{breiman2001} and support vector machines \cite{cortes1995} are two of the most established such learners, and we use the scikit-learn implementations \cite{pedregosa2011} of both in this work.

\textbf{What is missing.} No study we are aware of targets building setback violation detection from street-level imagery in a South Asian regulatory framework, and none directly compares raw-pixel, frozen-embedding, and end-to-end fine-tuned representations on the same regulatory classification task. This paper addresses both gaps using RAJUK's setback definitions \cite{rajuk2008,bnbc2020} and a dataset grounded in Dhaka's actual urban morphology.

\section{Dataset and Annotation}

The dataset consists of 372 labeled images collected by the authors through on-foot surveys of selected neighborhoods within RAJUK's jurisdiction, all captured using mobile phone cameras. The image dataset, including the annotation sheet used throughout this study, is publicly available.

\subsection{Study Area and Regulatory Context}

Our study covers buildings within the Dhaka Metropolitan Area, RAJUK's jurisdiction. The Dhaka Metropolitan Building Construction Rules (2008) \cite{rajuk2008} and the Bangladesh National Building Code (2020) \cite{bnbc2020} specify minimum setback distances from road boundaries, varying by road width and land-use zone. Under this framework, a building is considered a setback violator if its ground-floor structure, or a permanent extension such as a boundary wall, ramp, or covered veranda, encroaches on or beyond the mandated setback line.

\subsection{Image Acquisition}

All images were captured at a minimum resolution of 1920$\times$1080 pixels, held in landscape orientation, with the photographer standing on the opposite side of the road to obtain full facade visibility. We did not control for real-world conditions: images were taken at different times of day (353 daytime and 19 nighttime captures) under varied lighting (sunny, overcast, shaded) and with ordinary street obstructions such as parked vehicles, vendors, and pedestrians in frame, mirroring the conditions a deployed system would face. For each image we recorded the neighborhood, the time of capture, and two measured distances in feet: the building-to-boundary setback distance and the camera-to-facade distance. Recorded setback distances range from 0 to 12 feet and serve as a quantitative anchor for the categorical label.

\subsection{Annotation Protocol}

Rather than bounding-box annotation, we used a structured image-level protocol grounded directly in RAJUK's setback regulations. Each image received one of three labels:
\begin{itemize}
\item \textbf{Positive / violation:} the ground-floor structure, or a permanent extension such as a boundary wall, ramp, or covered veranda, is judged to encroach on or beyond the mandated setback space. 61 images.
\item \textbf{Negative / compliant:} a clear, unobstructed setback space is visible between the facade and the road or footpath boundary. 229 images.
\item \textbf{Moderate / borderline:} setback compliance cannot be determined with confidence from visual inspection alone, due to partial occlusion, a restricted viewing angle, poor lighting, roadside obstructions, or genuine ambiguity about the plot boundary. 82 images.
\end{itemize}

\begin{figure}[htbp]
\centering
\textbf{Negative (compliant)}\\[1mm]
\includegraphics[width=0.30\linewidth]{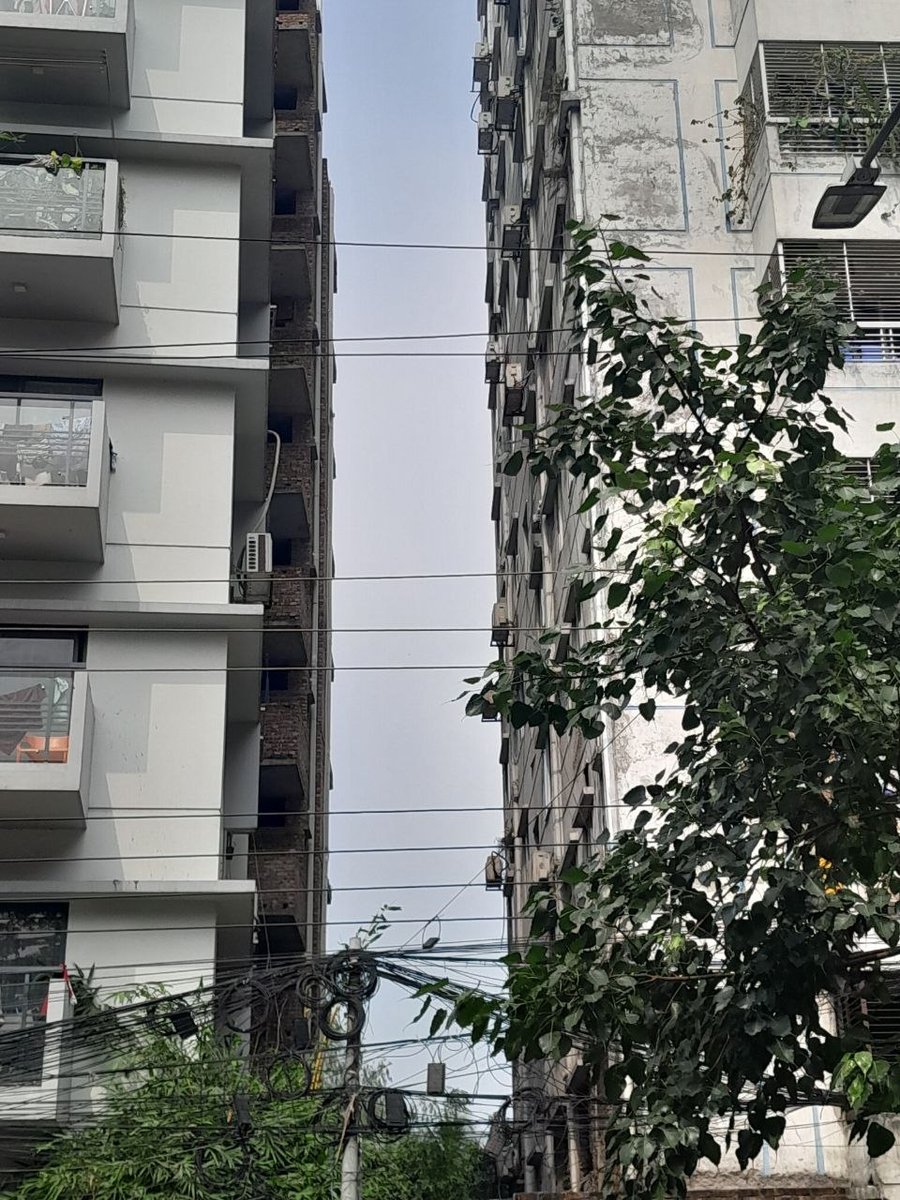}\hfill
\includegraphics[width=0.30\linewidth]{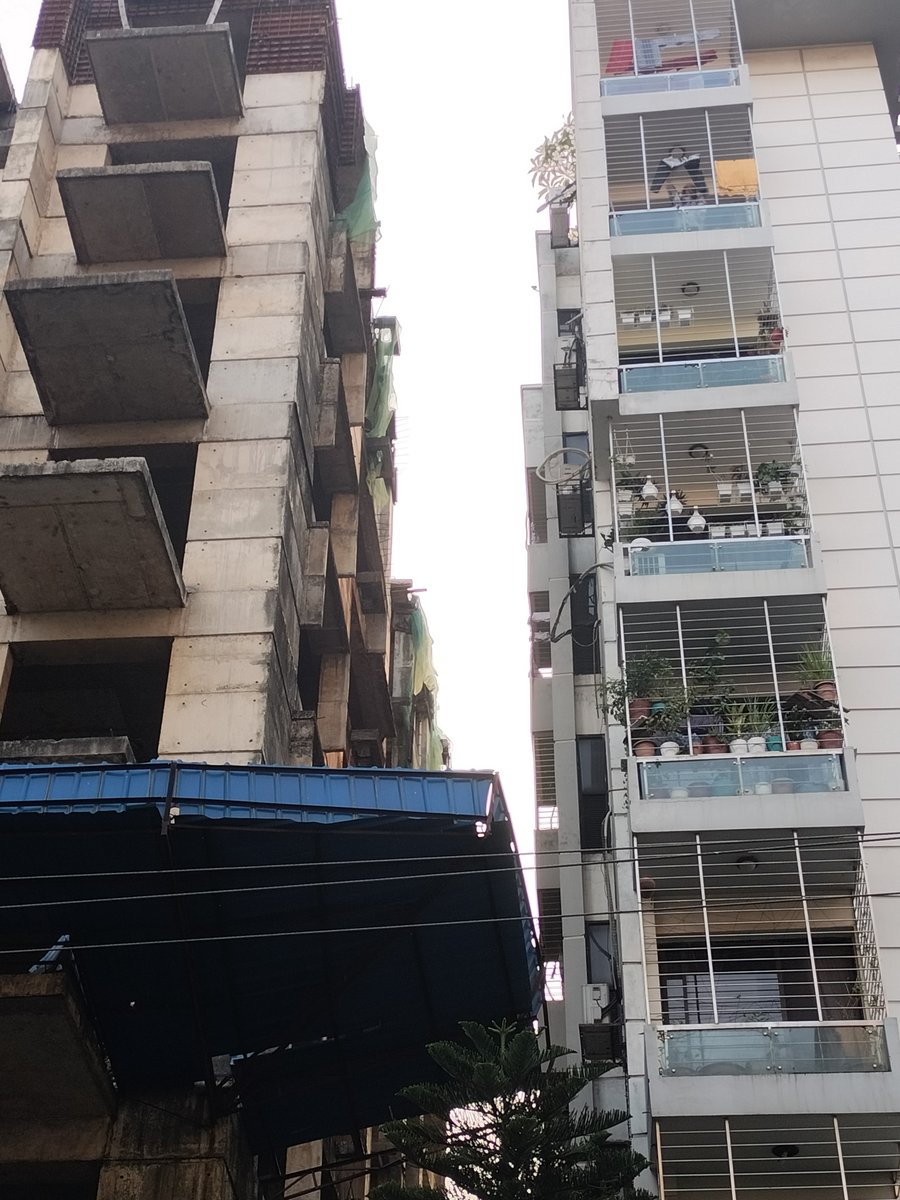}\hfill
\includegraphics[width=0.30\linewidth]{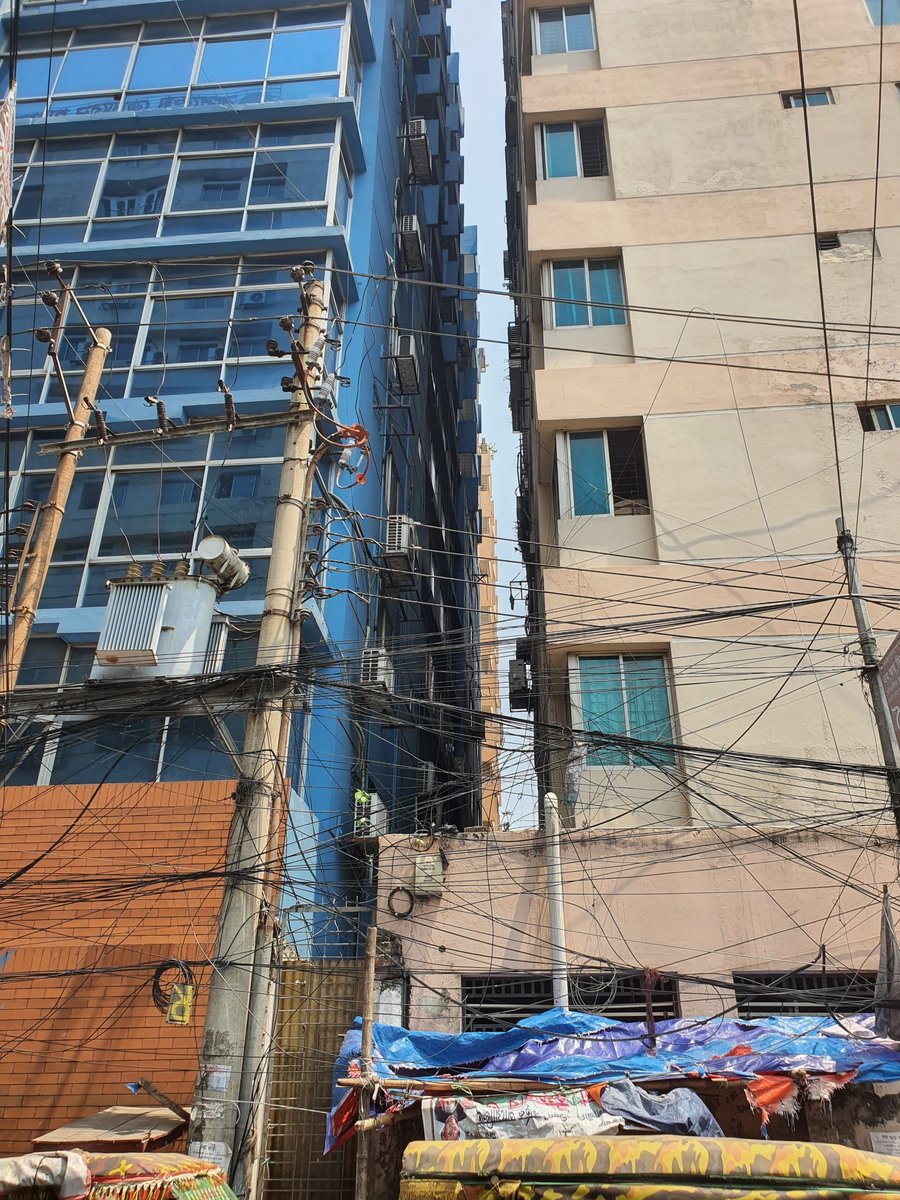}\\[3mm]
\textbf{Moderate (borderline)}\\[1mm]
\includegraphics[width=0.30\linewidth]{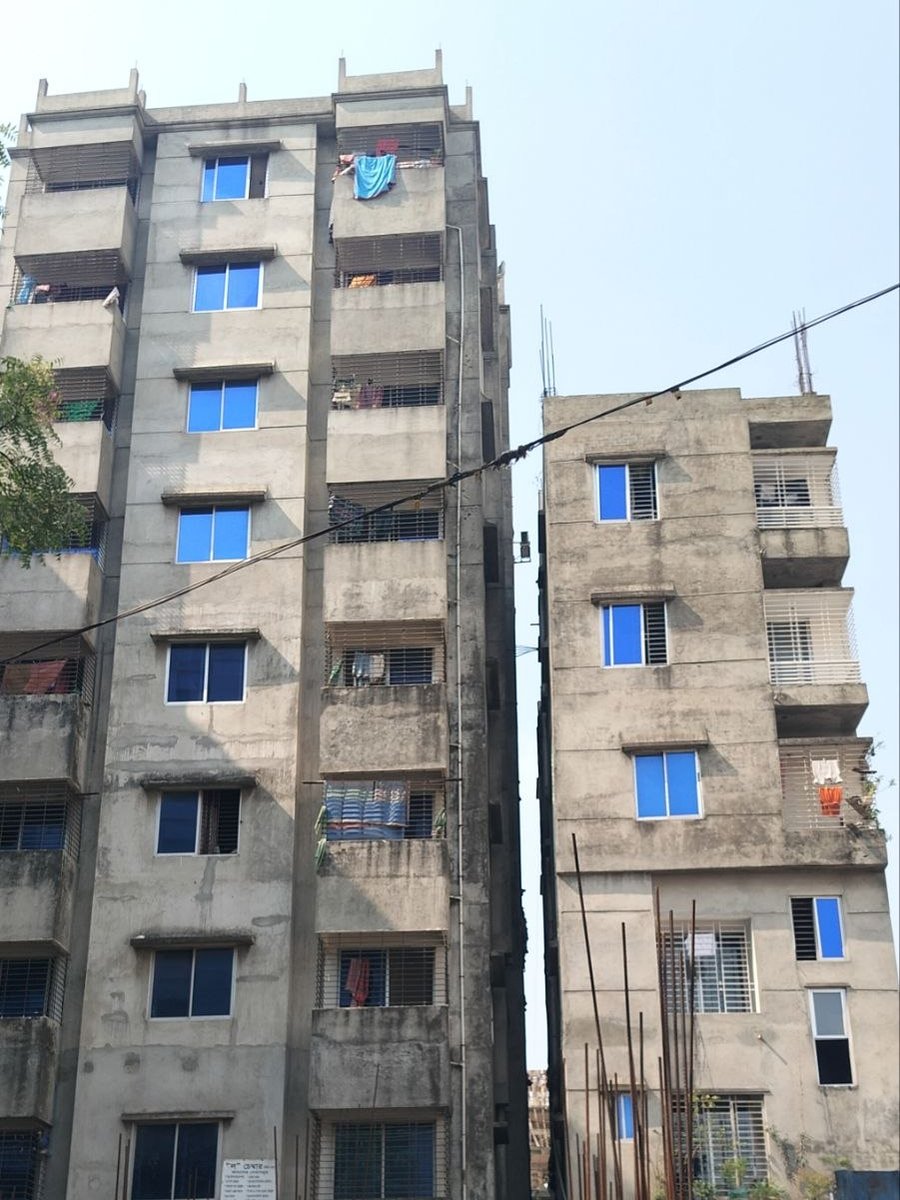}\hfill
\includegraphics[width=0.30\linewidth]{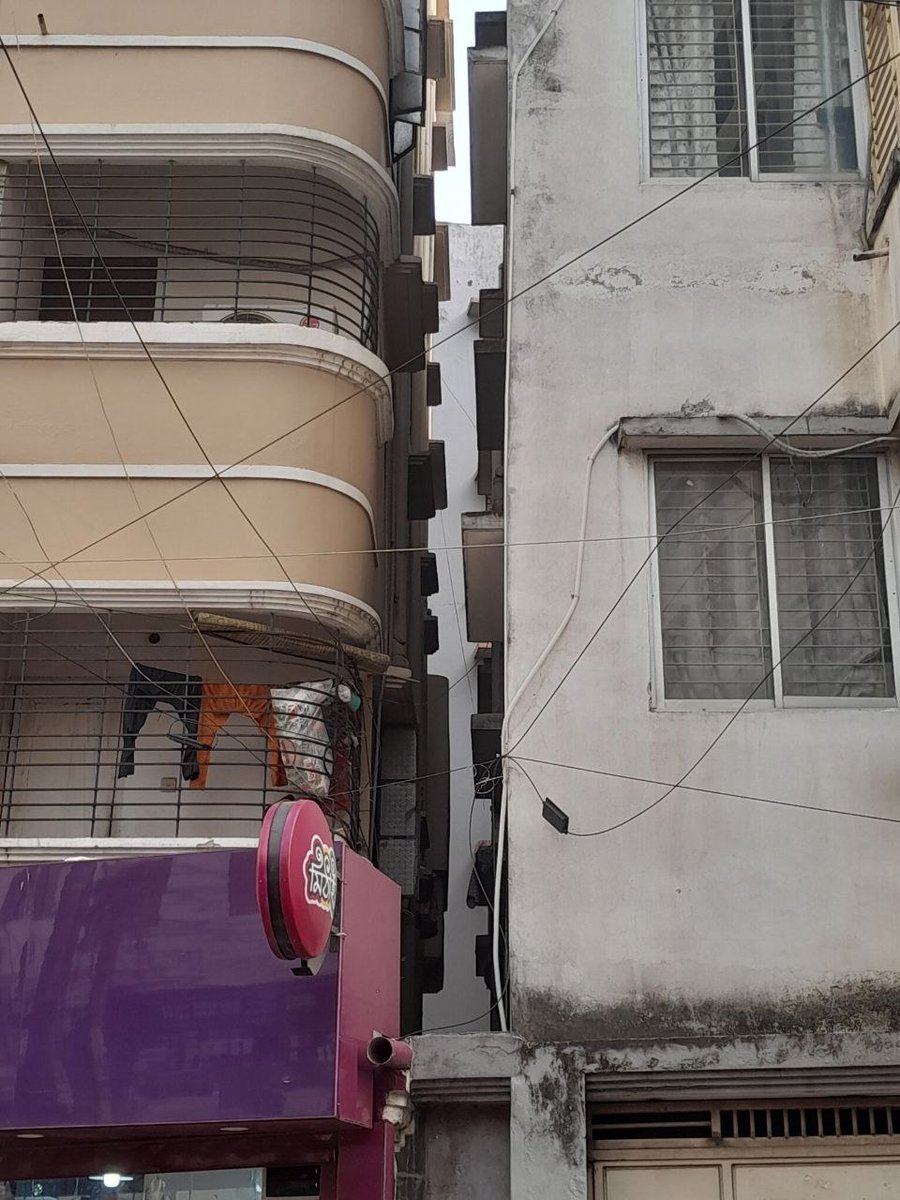}\hfill
\includegraphics[width=0.30\linewidth]{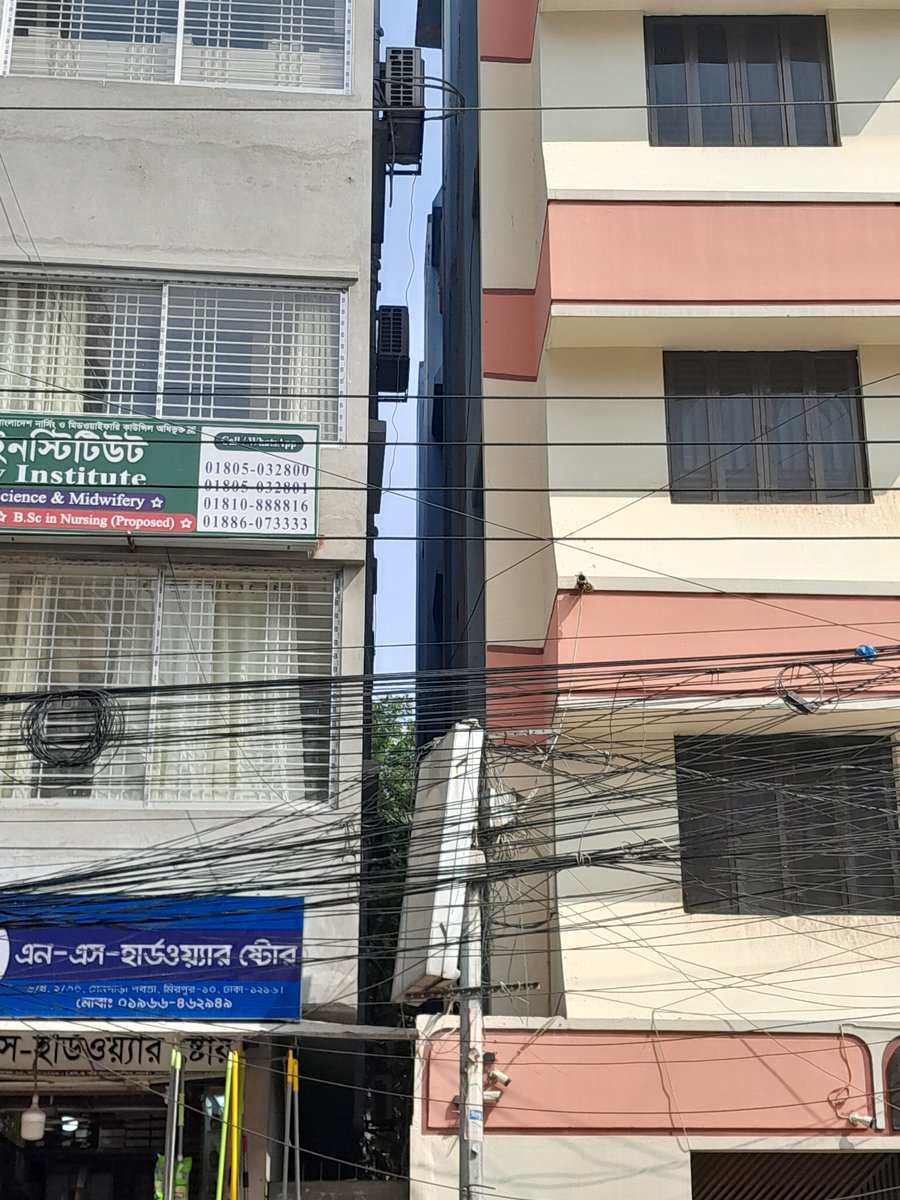}\\[3mm]
\textbf{Positive (violation)}\\[1mm]
\includegraphics[width=0.30\linewidth]{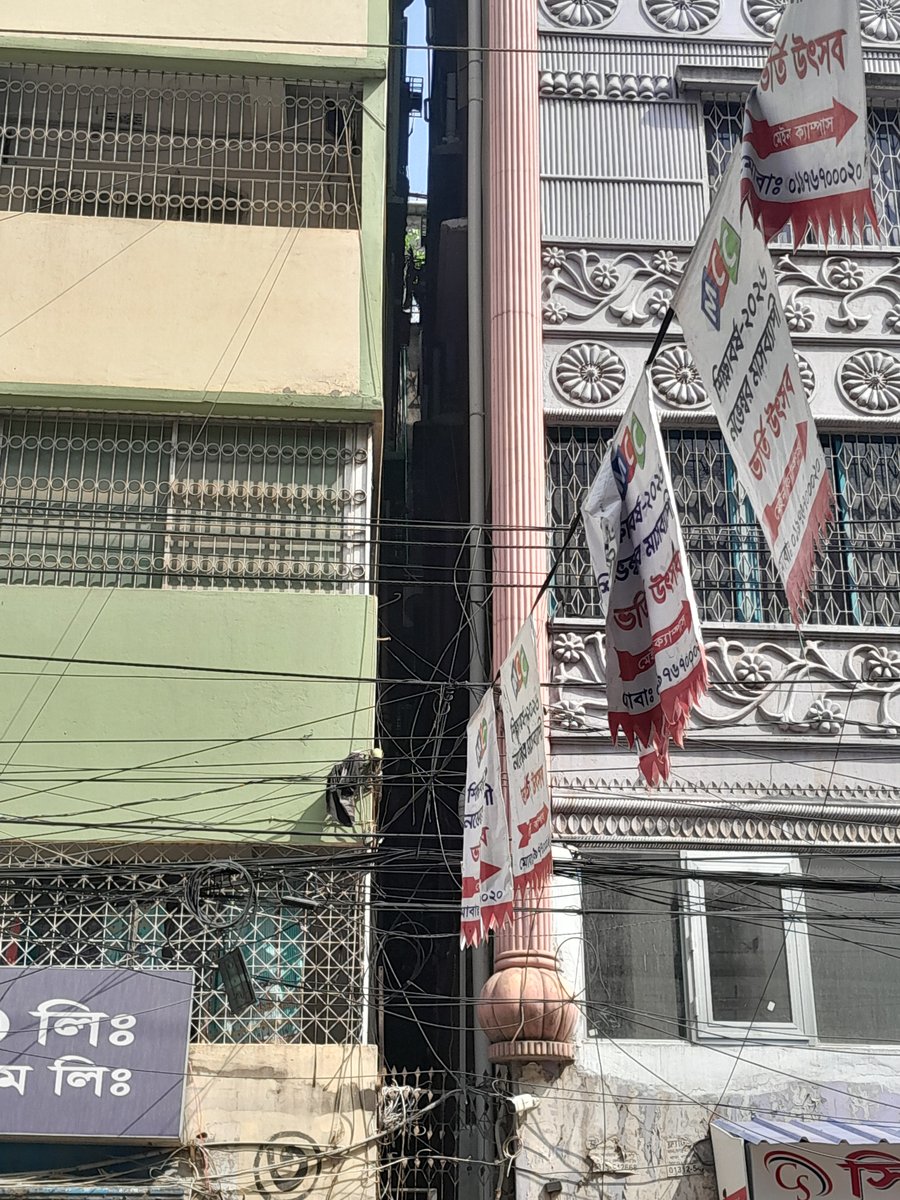}\hfill
\includegraphics[width=0.30\linewidth]{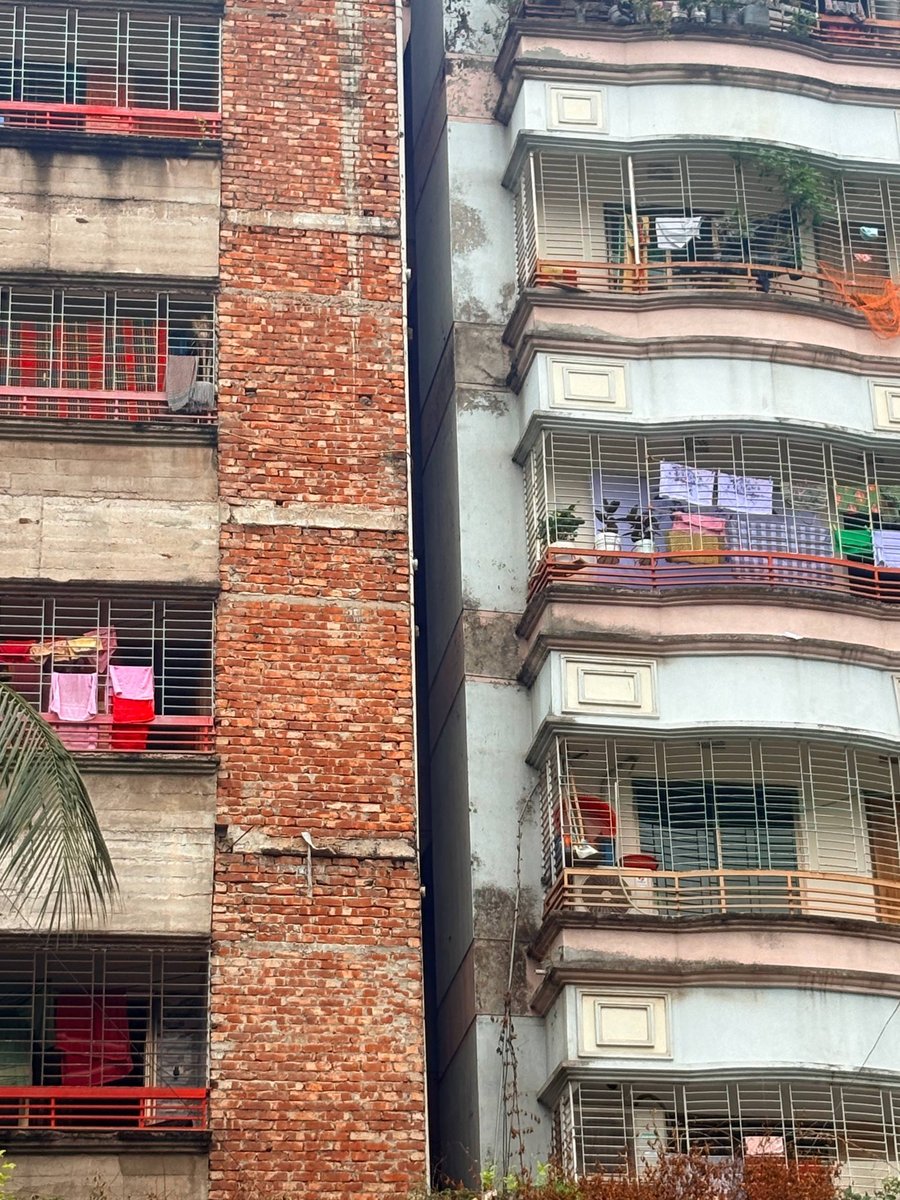}\hfill
\includegraphics[width=0.30\linewidth]{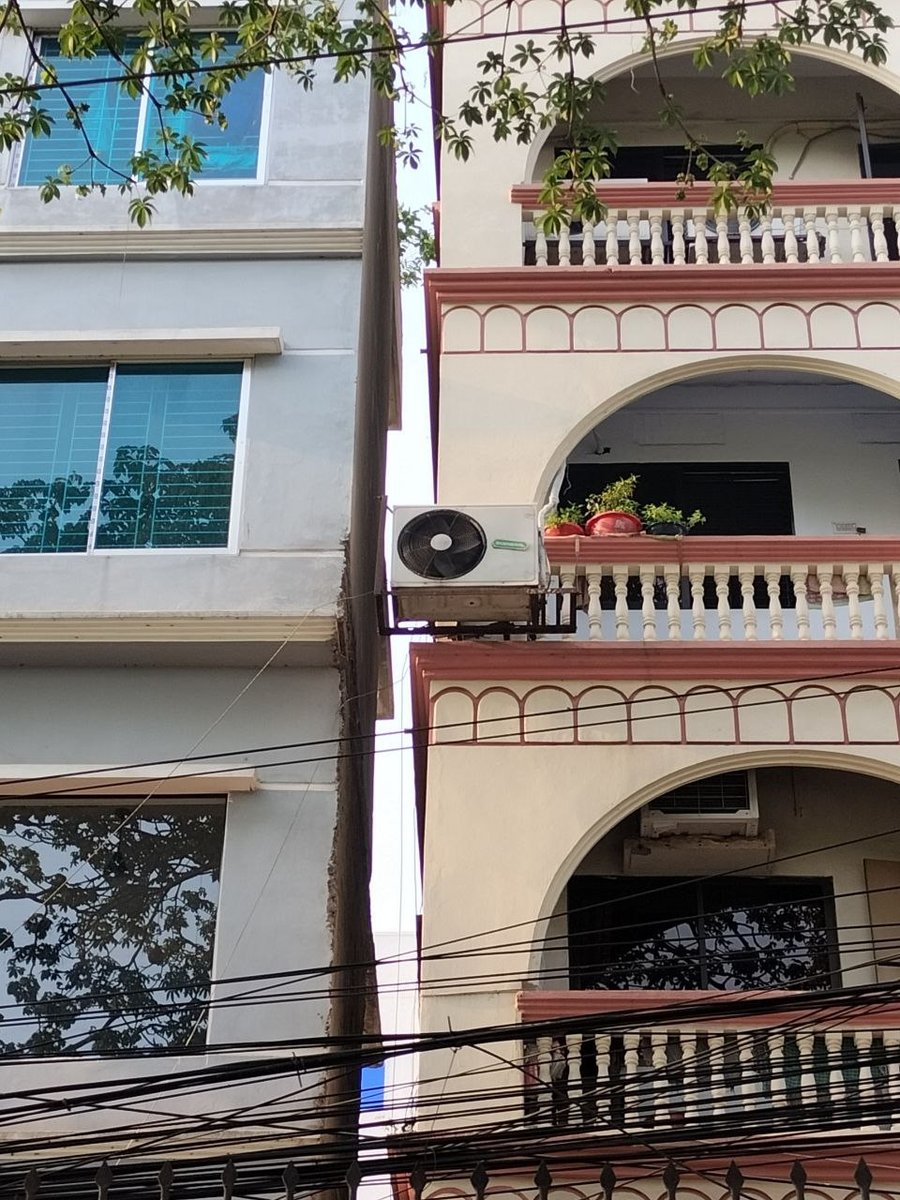}
\caption{Representative labeled images from dataset, one row per class. Negative examples show a visible, unobstructed setback; moderate examples are borderline or visually ambiguous; positive examples show an apparent setback violation.}
\label{fig:samples}
\end{figure}

Each image was reviewed by the authors against the visual evidence of setback space and the recorded distance measurement. Disagreement between reviewers, or insufficient visual clarity for a confident decision, resulted in the image being placed in the moderate class. Figure~\ref{fig:samples} shows one example from each class for three different buildings.

\subsection{Dataset Statistics}

Table~\ref{tab:stats} summarizes the dataset. The class distribution is skewed toward the compliant (negative) class, consistent with field observations, and we address this imbalance explicitly during training through per-class loss weighting or, for the SVM pipeline, exact class balancing (Section~\ref{sec:algorithms}). Table~\ref{tab:areas} gives the per-neighborhood breakdown; two images have no recorded neighborhood.

\begin{table}[htbp]
\centering
\small
\begin{tabularx}{\linewidth}{l X}
\toprule
\textbf{Attribute} & \textbf{Value} \\
\midrule
Total labeled images & 372 \\
Positive / violation & 61 \\
Negative / compliant & 229 \\
Moderate / borderline & 82 \\
Augmented images generated & 5,968 \\
Minimum resolution & 1920 $\times$ 1080 px \\
Capture device & Smartphone cameras \\
Recorded metadata & Building-to-boundary distance and camera distance (feet) \\
Study area & RAJUK jurisdiction, Dhaka Metropolitan Area \\
Neighborhoods covered & 16 (Table~\ref{tab:areas}) \\
Collection time & Day: 353, Night: 19 \\
Collection period & 2026 \\
\bottomrule
\end{tabularx}
\caption{Summary statistics of the dataset.}
\label{tab:stats}
\end{table}

\begin{table}[htbp]
\centering
\footnotesize
\begin{tabularx}{0.75\linewidth}{X r}
\toprule
\textbf{Neighborhood} & \textbf{Images} \\
\midrule
Khilgaon & 98 \\
Mirpur 10 & 59 \\
Monsurabad & 44 \\
Mohammadpur & 48 \\
Changkharpul & 29 \\
Dhanmondi & 28 \\
Wari & 15 \\
Mirpur 1 & 9 \\
Bashundhara & 8 \\
Lalbag & 7 \\
Purana Paltan & 7 \\
Mirpur 12 & 5 \\
Shyamoli & 6 \\
Segunbagicha & 3 \\
Puran Dhaka & 3 \\
Bosila & 1 \\
Unrecorded & 2 \\
\midrule
Total & 372 \\
\bottomrule
\end{tabularx}
\caption{Number of labeled images per neighborhood.}
\label{tab:areas}
\end{table}

Of the 373 photographs captured in the field, 356 could be matched to a unique subject identifier with a usable label and image file; these 356 subjects form the basis of the leakage-free splitting procedure described in Section~\ref{sec:split}.

\section{Learning Algorithms}
\label{sec:algorithms}

We compare four models that assign each image to one of three classes: negative, moderate, or positive. They differ in how the image is represented before classification: raw pixels (MLP), frozen convolutional embeddings passed to classical ensembles (Random Forest ensemble, balanced SVM), or convolutional features learned end to end (CNN).

\subsection{Leakage-Free Subject-Level Splitting}
\label{sec:split}

Because each of the 356 real subjects has multiple augmented copies, a naive random or stratified split at the image level would place near-duplicate views of the same building in both the training and evaluation sets, inflating measured accuracy. We instead split at the level of unique subject identifiers before any augmented image is assigned to a partition: real-image identifiers are partitioned into training, validation, and test subsets by stratified sampling, and augmented copies are then assigned exclusively to whichever partition their parent subject was assigned to.

The Random Forest ensemble, the balanced SVM ensemble, and the initial split used by the CNN pipeline share a common subject-level partition targeting a 70:20:10 train:test:validation ratio over the 356 real subjects, giving 4,420 training images (782 real plus augmented, class counts 2,720 negative, 969 moderate, 731 positive), 646 validation images (408/136/102), and 1,275 test images (782/289/204). The CNN pipeline additionally applies a hard-example mining step described in Section~\ref{sec:cnn} that moves some validation images into training; the test set is left untouched by this step, so the CNN, Random Forest, and SVM results in Section~\ref{sec:results} are all measured on the identical 1,275-image test set. The MLP baseline uses its own, independently constructed subject-level split, reserving only 56 real test images (Section~\ref{sec:mlp}), and its results are therefore reported separately rather than compared image-for-image against the other three models.

Per-class loss weights, where a model uses weighted cross-entropy, are computed as the normalized inverse of class frequency in the relevant training split,
\begin{equation}
w_c = \frac{1/n_c}{\sum_{c'} 1/n_{c'}},
\end{equation}
where $n_c$ is the number of training images in class $c$. On the shared 4,420-image training split this gives weights of approximately 0.133 (negative), 0.373 (moderate), and 0.494 (positive), reflecting the relative scarcity of borderline and violation images.

\subsection{Multi-Layer Perceptron Baseline}
\label{sec:mlp}

The MLP is a fully connected network with no convolutional or attention component, intended to measure how much can be achieved from raw pixel statistics alone, without the spatial inductive bias of a convolutional architecture such as VGG \cite{simonyan2014}, ResNet \cite{he2016}, or DenseNet \cite{huang2017}. Each image is resized to 128$\times$128 pixels and flattened to a vector of length $128 \times 128 \times 3 = 49{,}152$. Four hidden blocks of decreasing width apply a linear layer, one-dimensional batch normalization, ReLU activation, and dropout (rate 0.40), followed by a linear output layer:
\[
49{,}152 \rightarrow 2{,}048 \rightarrow 1{,}024 \rightarrow 512 \rightarrow 256 \rightarrow 3.
\]
The network has 103,428,099 trainable parameters. Its own subject-level split gives 4,420 training images, 57 validation images, and 56 test images, all real. Table~\ref{tab:mlp-hparams} summarizes training configuration.

\begin{table}[htbp]
\centering
\small
\begin{tabularx}{\linewidth}{l X}
\toprule
\textbf{Parameter} & \textbf{Value} \\
\midrule
Input dimension & 49,152 (128 $\times$ 128 $\times$ 3, flattened) \\
Hidden layer widths & 2048, 1024, 512, 256 \\
Activation / normalization & ReLU / one-dimensional batch normalization \\
Dropout rate & 0.40 \\
Total parameters & 103,428,099 \\
Loss function & Weighted cross-entropy \\
Optimizer & AdamW, learning rate $10^{-3}$ (fine-tuning stage $2\times10^{-4}$) \\
Learning rate schedule & Cosine annealing, $10^{-3} \rightarrow 10^{-6}$ \\
Batch size & 32 \\
Max epochs / patience & 60 / 10 \\
Cross-validation & Stratified 5-fold, real images only \\
Augmentations (train only) & Horizontal flip ($p=0.5$), vertical flip ($p=0.3$), brightness/contrast/color jitter; no spatial transforms, since a flat MLP has no notion of spatial structure \\
Hardware & NVIDIA T4 GPU \\
\bottomrule
\end{tabularx}
\caption{Training configuration for the MLP baseline.}
\label{tab:mlp-hparams}
\end{table}

Stratified 5-fold cross-validation, run only on the 260 real training images (208 train / 52 validation per fold) so that augmented near-duplicates cannot leak across a fold boundary, gave per-fold accuracies of 0.6154, 0.5577, 0.5192, 0.6154, and 0.6154, a mean of 0.5846 $\pm$ 0.0396. The best cross-validation checkpoint was then fine-tuned on the full 4,420-image training split for five epochs at a reduced learning rate. Training accuracy rose steadily across fine-tuning, reaching 89.3\% by the last epoch, while validation accuracy reached only 42.1\% (loss 2.098) and test accuracy reached 50.0\% (loss 1.852), a clear sign of overfitting on a small evaluation set. Because the reported training accuracy is measured under live data augmentation at the final fine-tuning epoch rather than through a deterministic re-evaluation, it should be read as an upper bound on memorization rather than a figure directly comparable to the training accuracy of the other three models.

Table~\ref{tab:mlp-perclass} gives the test-set classification report. The model classifies the majority negative class most reliably; the moderate class is over-predicted, giving high recall but low precision, and the positive class is rarely recovered, consistent with the class imbalance and the model's reliance on coarse pixel statistics rather than spatial structure.

\begin{table}[htbp]
\centering
\small
\begin{tabular}{lcccc}
\toprule
\textbf{Class} & \textbf{Precision} & \textbf{Recall} & \textbf{F1} & \textbf{Support} \\
\midrule
Negative & 0.76 & 0.54 & 0.63 & 35 \\
Moderate & 0.29 & 0.58 & 0.39 & 12 \\
Positive & 0.29 & 0.22 & 0.25 & 9 \\
\bottomrule
\end{tabular}
\caption{MLP test-set classification report ($n=56$).}
\label{tab:mlp-perclass}
\end{table}

\begin{figure}[htbp]
\centering
\includegraphics[width=\linewidth]{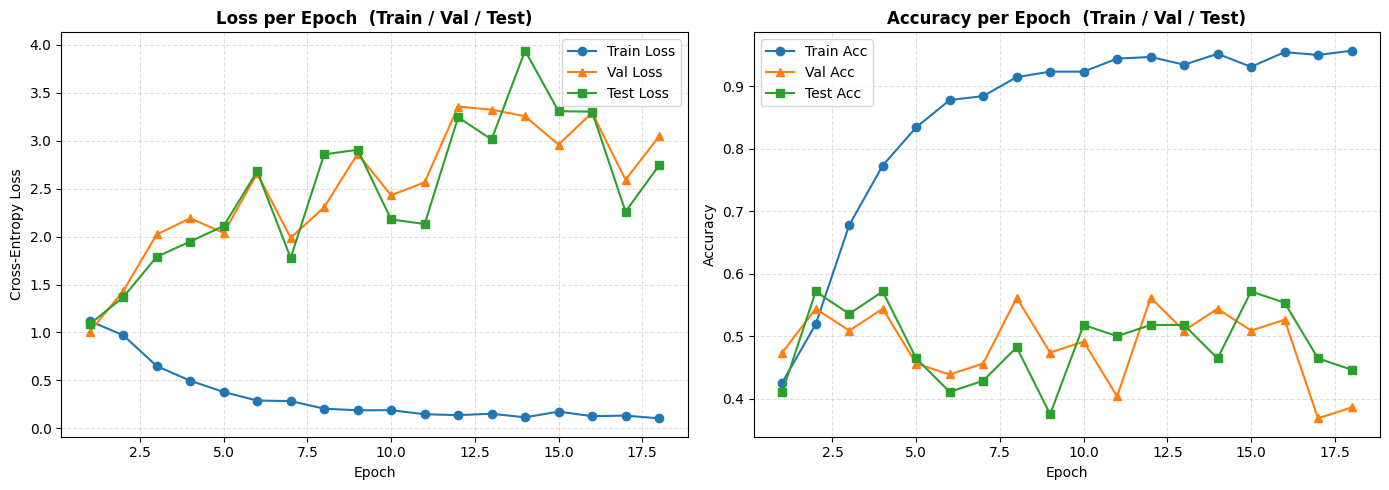}
\caption{MLP loss and accuracy per epoch (train / validation / test) during full-split training.}
\label{fig:mlp-curves}
\end{figure}

\begin{figure}[htbp]
\centering
\includegraphics[width=0.75\linewidth]{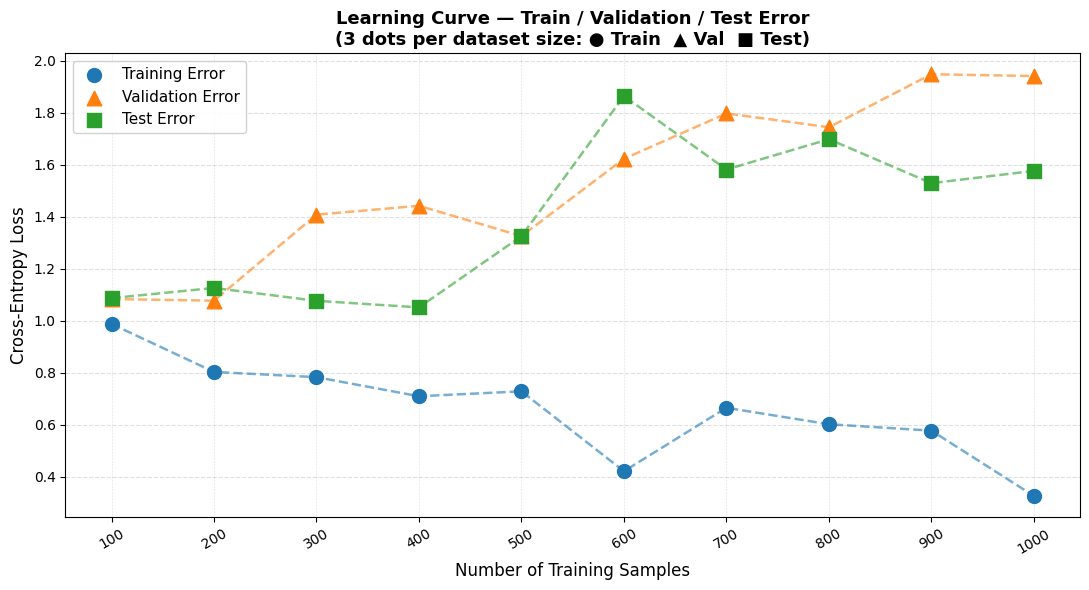}
\caption{MLP learning curve: cross-entropy loss for training, validation, and test sets as training set size grows from 100 to 1,000 samples.}
\label{fig:mlp-lc}
\end{figure}

Figure~\ref{fig:mlp-curves} shows per-epoch loss and accuracy during full-split training, and Figure~\ref{fig:mlp-lc} shows the corresponding learning curve as training set size grows. Training loss falls steadily while validation and test loss diverge upward, and additional data alone does not close the gap, which is consistent with a representation (raw pixels) that lacks the inductive bias needed for this task.

\subsection{Random Forest Heterogeneous Ensemble}
\label{sec:rf}

The second model is a heterogeneous soft-voting ensemble built on frozen 2,048-dimensional, L2-normalized embeddings from an ImageNet-pretrained ResNet50 backbone (frozen; only the classifier stage is trained). Four base learners are trained independently on the shared 4,420-image training split:
\begin{itemize}
\item a bootstrapped Random Forest \cite{breiman2001} (750 trees, entropy criterion, \texttt{max\_features}=$\sqrt{\cdot}$, bootstrap sample fraction 0.90, per-class weights $\{1.0, 2.15, 2.35\}$);
\item a bootstrapped Extra Trees ensemble with the same tree count and sampling settings;
\item a balanced multinomial logistic regression ($C=0.20$, L-BFGS solver);
\item a balanced RBF-kernel support vector machine \cite{cortes1995} ($C=1.5$), included in the search space but not necessarily in the final ensemble.
\end{itemize}
All four are wrapped with median imputation and standardization where applicable and trained using scikit-learn \cite{pedregosa2011}.

Soft-voting weights and mild per-class probability multipliers are selected on the validation split by a randomized search over roughly 9,600 weight/multiplier combinations, scored by a composite of image-level accuracy, subject-level accuracy, real-image accuracy, subject-level macro-F1, and subject-level balanced accuracy. The search selected weights of 0.015 (Random Forest), 0.590 (Extra Trees), 0.388 (logistic regression), and 0.007 (RBF-SVM), with class multipliers 1.0 (negative), 1.2 (moderate), and 1.4 (positive) applied to the blended probabilities. Because the SVM's selected weight fell below a 0.01 inclusion threshold, the deployed ensemble is effectively three learners, Random Forest, Extra Trees, and logistic regression, dominated by Extra Trees. The selected configuration is then refit on the combined training and validation data (5,066 images) and evaluated once on the untouched 1,275-image test set.

\begin{figure}[htbp]
\centering
\includegraphics[width=\linewidth]{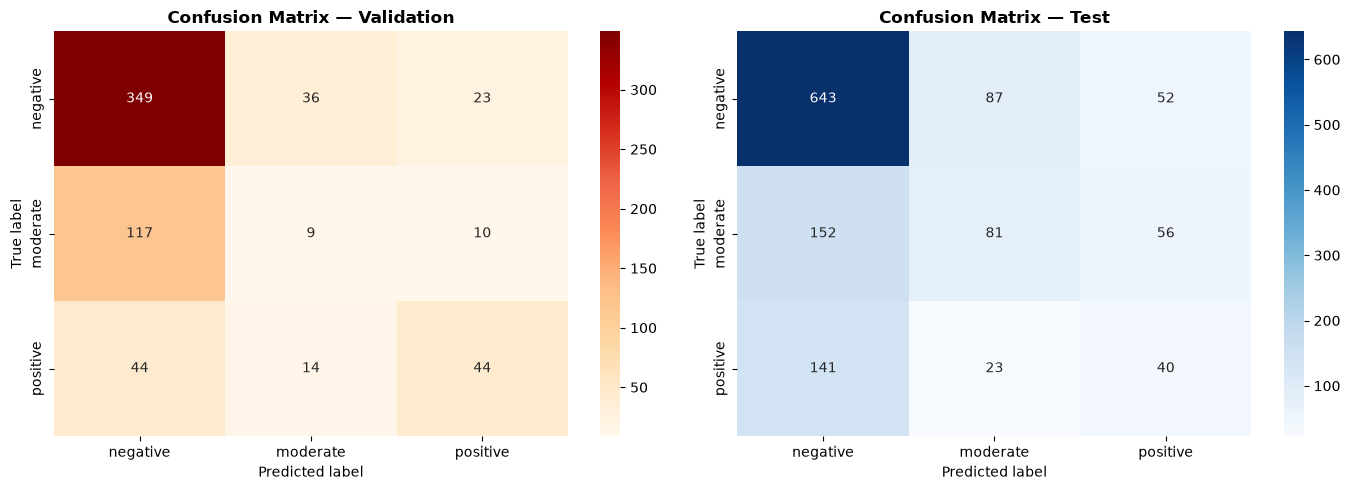}
\caption{Random Forest ensemble confusion matrices on the validation set (left, $n=646$) and the held-out test set (right, $n=1{,}275$).}
\label{fig:rf-cm}
\end{figure}

The final ensemble reaches 100.0\% accuracy on its own refit training data, 62.2\% accuracy on the validation split used for model selection, and 59.9\% accuracy (log-loss 0.883) on the held-out test set, with a test macro-F1 of 0.438. Table~\ref{tab:rf-perclass} gives the per-class breakdown, computed from the confusion matrix in Figure~\ref{fig:rf-cm}.

\begin{table}[htbp]
\centering
\small
\begin{tabular}{lcccc}
\toprule
\textbf{Class} & \textbf{Precision} & \textbf{Recall} & \textbf{F1} & \textbf{Support} \\
\midrule
Negative & 0.69 & 0.82 & 0.75 & 782 \\
Moderate & 0.42 & 0.28 & 0.34 & 289 \\
Positive & 0.27 & 0.20 & 0.23 & 204 \\
\bottomrule
\end{tabular}
\caption{Random Forest ensemble test-set classification report ($n=1{,}275$).}
\label{tab:rf-perclass}
\end{table}

\begin{figure}[htbp]
\centering
\includegraphics[width=\linewidth]{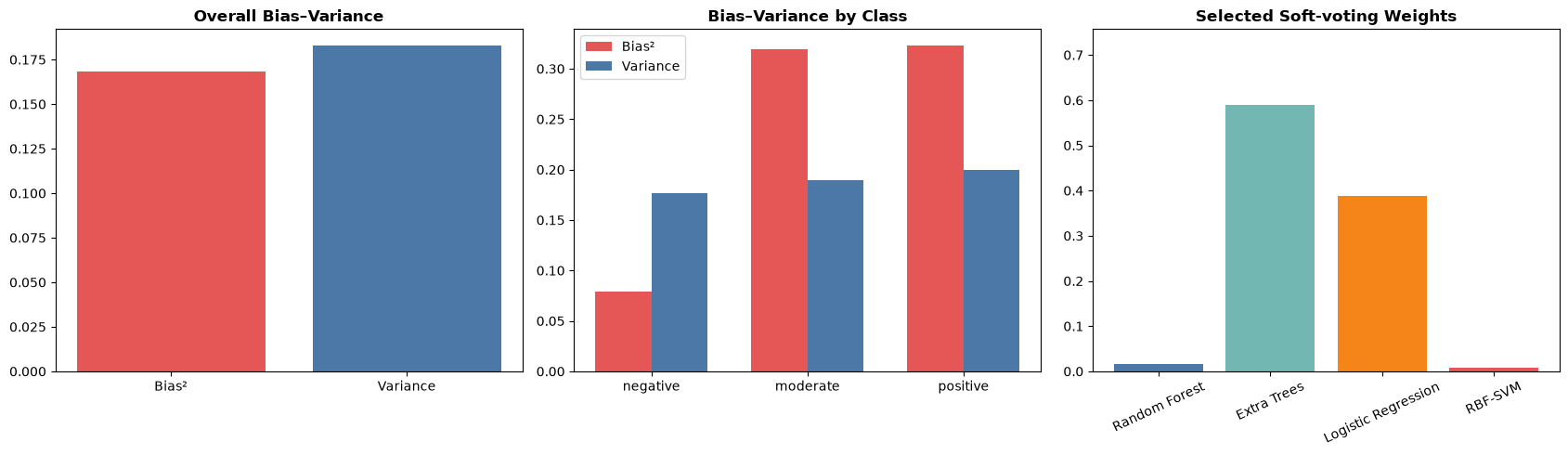}
\caption{Random Forest ensemble bias-variance decomposition on the validation set, overall (left), by class (middle), and the selected soft-voting weights (right).}
\label{fig:rf-bv}
\end{figure}

Figure~\ref{fig:rf-bv} decomposes validation error into bias and variance for the tree-based ensemble members. Bias and variance are of comparable, moderate magnitude overall, but both are markedly higher for the moderate and positive classes than for the negative class, indicating that the ensemble both under-fits and is inconsistent across bootstrap samples specifically on the minority classes.

\subsection{Balanced Bagged Support Vector Machine}
\label{sec:svm}

The third model is a bagging ensemble of linear support vector machines \cite{cortes1995}, trained on frozen 1,536-dimensional embeddings formed by concatenating average- and max-pooled activations from the \texttt{layer3} and \texttt{layer4} blocks of an ImageNet-pretrained ResNet18, L2-normalized. Unlike the other three models, class imbalance is addressed here through exact resampling rather than loss weighting: for both model selection and the final refit, each class is represented by an identical number of images (up to 700 per class, at most 40 real images per class supplemented with randomly selected augmented copies), so every training bag sees a perfectly balanced label distribution.

Each ensemble member is a custom GPU-resident multiclass hinge-loss linear classifier, optimized with AdamW directly on the embeddings; bagging combines 15 to 17 such members with per-class-balanced bootstrap sampling and random feature subsampling. A small grid of four configurations, varying the regularization strength $C$ from 0.01 to 0.10 together with the bagging sample and feature fractions, was evaluated on the validation split; the best single configuration reached a validation macro-F1 of 0.585, and blending the top three configurations by a further weighted search raised validation macro-F1 to 0.519 at the selected blend (0.75 weight on the best single configuration, 0.25 on a second). Table~\ref{tab:svm-hparams} summarizes the configuration space and the selected setup.

\begin{table}[htbp]
\centering
\small
\begin{tabularx}{\linewidth}{l X}
\toprule
\textbf{Parameter} & \textbf{Value} \\
\midrule
Feature representation & ResNet18 \texttt{layer3}+\texttt{layer4} pooled embedding, 1,536-D, L2-normalized \\
Base learner & Linear multiclass hinge-loss SVM, AdamW optimized on GPU \\
Ensemble structure & Bagging, 15--17 members, per-class-balanced bootstrap sampling, random feature subsampling \\
Regularization $C$ (search range) & 0.01, 0.03, 0.05, 0.10 \\
Class balancing & Exact per-class resampling to 700 images/class (up to 40 real, remainder augmented) \\
Model selection & Weighted blend of top-3 configurations by validation macro-F1 and balanced accuracy \\
Hardware & NVIDIA GeForce GTX 1650, PyTorch CUDA 12.4, mixed precision (fp16) \\
\bottomrule
\end{tabularx}
\caption{Configuration of the balanced bagged SVM ensemble.}
\label{tab:svm-hparams}
\end{table}

\begin{figure}[htbp]
\centering
\includegraphics[width=\linewidth]{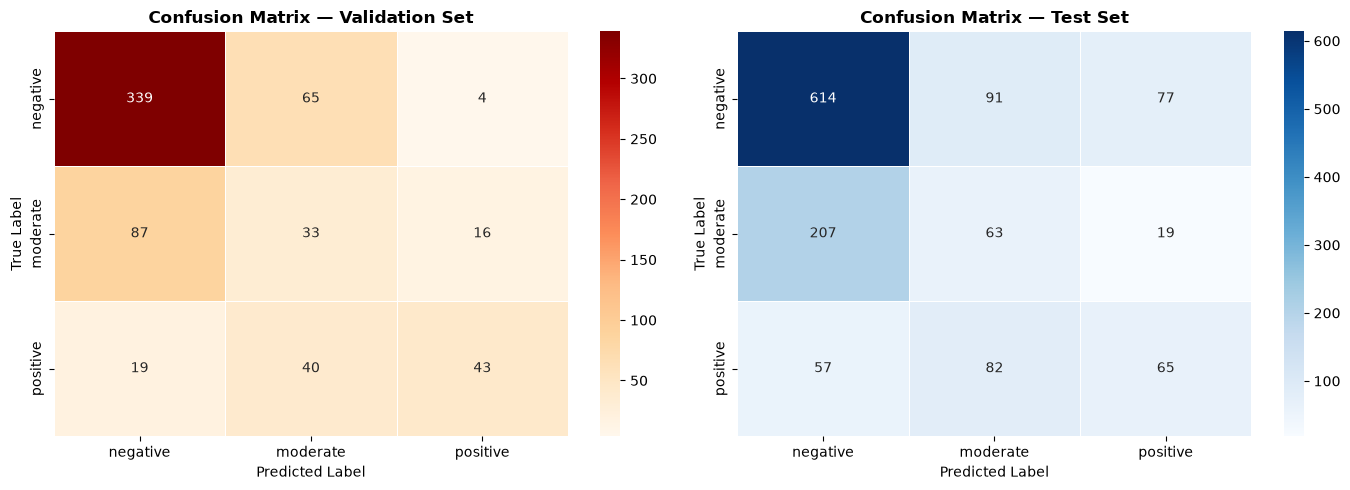}
\caption{Balanced SVM ensemble confusion matrices on the validation set (left, $n=646$) and the held-out test set (right, $n=1{,}275$).}
\label{fig:svm-cm}
\end{figure}

The final refit reaches 100.0\% accuracy on its own balanced training pool (2,100 images, loss 0.337), 64.2\% accuracy on the validation split used for selection (loss 0.755), and 58.2\% accuracy (loss 0.921) on the same 1,275-image held-out test set used for the Random Forest ensemble and the CNN, with a test macro-F1 of 0.445. Table~\ref{tab:svm-perclass} and Figure~\ref{fig:svm-cm} give the per-class breakdown and confusion matrices.

\begin{table}[htbp]
\centering
\small
\begin{tabular}{lcccc}
\toprule
\textbf{Class} & \textbf{Precision} & \textbf{Recall} & \textbf{F1} & \textbf{Support} \\
\midrule
Negative & 0.70 & 0.79 & 0.74 & 782 \\
Moderate & 0.27 & 0.22 & 0.24 & 289 \\
Positive & 0.40 & 0.32 & 0.36 & 204 \\
\bottomrule
\end{tabular}
\caption{Balanced SVM ensemble test-set classification report ($n=1{,}275$).}
\label{tab:svm-perclass}
\end{table}

\begin{figure}[htbp]
\centering
\includegraphics[width=0.85\linewidth]{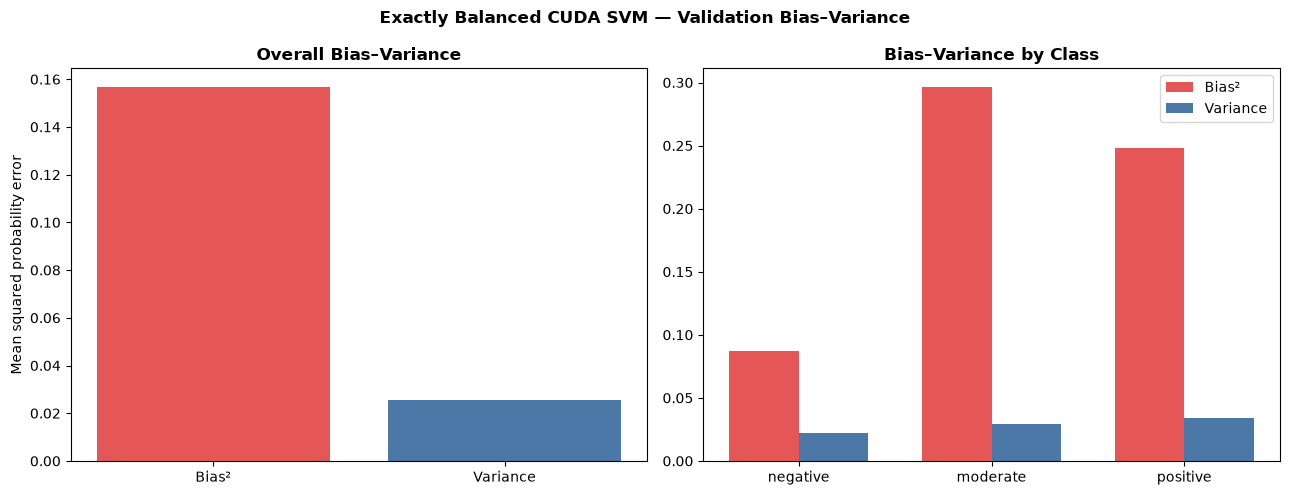}
\caption{Balanced SVM ensemble bias-variance decomposition on the validation set, overall (left) and by class (right).}
\label{fig:svm-bv}
\end{figure}

Figure~\ref{fig:svm-bv} shows the corresponding bias-variance decomposition. In contrast with the Random Forest ensemble, variance is small relative to bias for every class, and bias is highest for the moderate and positive classes; the exact class-balancing strategy appears to trade some of the ensemble's flexibility for consistency, at the cost of a persistent, systematic error on the minority classes.

\subsection{Convolutional Neural Network: ResNet18 Transfer Learning}
\label{sec:cnn}

The fourth model fine-tunes a ResNet18 backbone, initialized with ImageNet-pretrained weights, end to end rather than treating it as a fixed feature extractor. With only 356 unique real subjects, training a convolutional network from scratch would overfit severely; transfer learning instead lets the network reuse general-purpose low-level detectors already learned from large-scale data. All layers except the final residual block (\texttt{layer4}) and the classification head are frozen, so 8,395,267 of the 11,178,051 total parameters (75.1\%) are trainable. The original fully connected layer is replaced with dropout (rate 0.5) followed by a linear layer to three classes. Images are resized to 224$\times$224 and normalized with ImageNet statistics.

Starting from the shared 70:20:10 split (Section~\ref{sec:split}), we apply a hard-example mining step: every validation image misclassified by a prior training run is identified (198 rows: 12 real, 186 augmented), its real copy is moved into training unconditionally, and roughly half of that subject's misclassified augmented copies are also moved in, so the model is explicitly re-exposed to the conditions it previously struggled with. This moves 107 images from validation into training and leaves the test set untouched. The resulting split is 4,527 training images (71.4\%), 539 validation images (8.5\%), and 1,275 test images (20.1\%, unchanged). Table~\ref{tab:cnn-hparams} summarizes the training configuration.

\begin{table}[htbp]
\centering
\small
\begin{tabularx}{\linewidth}{l X}
\toprule
\textbf{Parameter} & \textbf{Value} \\
\midrule
Backbone & ResNet18, ImageNet pretrained \\
Frozen layers & All except \texttt{layer4} and the classification head \\
Trainable parameters & 8,395,267 of 11,178,051 (75.1\%) \\
Input size & 224 $\times$ 224 $\times$ 3, ImageNet normalization \\
Classifier head & Dropout (0.5) followed by a linear layer to 3 classes \\
Loss function & Weighted cross-entropy with label smoothing (0.1) \\
Optimizer & AdamW, learning rate $3\times10^{-4}$ (fine-tuning stage $6\times10^{-5}$), weight decay $10^{-3}$ \\
Learning rate schedule & Cosine annealing, $T_{max}=25$ \\
Batch size & 16 \\
Max epochs / patience & 25 / 6 (fine-tuning stage: 8 / 3) \\
Augmentations (train only) & Random horizontal flip, random rotation ($\pm10^\circ$), color jitter, random resized crop (scale 0.8--1.0), Gaussian blur \\
Cross-validation & Stratified, subject-grouped 3-fold ($k=3$) \\
Hard-example mining & Misclassified validation subjects re-injected into training \\
Hardware & CUDA GPU \\
\bottomrule
\end{tabularx}
\caption{Training configuration for the CNN (ResNet18 transfer learning with hard-example mining).}
\label{tab:cnn-hparams}
\end{table}

Cross-validation here groups folds by subject identifier rather than by image, so augmented copies of a given building never appear in both the training and validation portions of the same fold; this is a stricter protocol than a subject-blind stratified fold, and it yields a correspondingly more conservative estimate. Three-fold cross-validation over the post-mining training split gave fold accuracies of 0.653, 0.632, and 0.651, a mean of 0.645 $\pm$ 0.010. The best cross-validation checkpoint was then fine-tuned for a further eight epochs at a reduced learning rate.

Table~\ref{tab:cnn-final} reports the final train, validation, and test error, all measured with the same deterministic resize-and-normalize transform (that is, without training-time augmentation), so the three figures are directly comparable to one another and to the other three models in this study. The model fully memorizes its training split (0.00\% error) while validation error is 10.0\% and test error is 34.6\%; the substantial gap between validation and test error, despite both coming from held-out subjects, reflects the fact that hard-example mining specifically removed the harder validation cases into training, leaving an easier residual validation set while the test set retains its original difficulty.

\begin{table}[htbp]
\centering
\small
\begin{tabular}{lcccc}
\toprule
\textbf{Split} & \textbf{Samples} & \textbf{Loss} & \textbf{Accuracy} & \textbf{Error rate} \\
\midrule
Training & 4,527 & 0.519 & 100.00\% & 0.00\% \\
Validation & 539 & 0.788 & 89.98\% & 10.02\% \\
Test & 1,275 & 1.025 & 65.41\% & 34.59\% \\
\bottomrule
\end{tabular}
\caption{Final train, validation, and test error for the fine-tuned CNN, evaluated with a deterministic, non-augmented transform on all three splits.}
\label{tab:cnn-final}
\end{table}

\begin{table}[htbp]
\centering
\small
\begin{tabular}{lcccc}
\toprule
\textbf{Class} & \textbf{Precision} & \textbf{Recall} & \textbf{F1} & \textbf{Support} \\
\midrule
Negative & 0.73 & 0.88 & 0.80 & 782 \\
Moderate & 0.55 & 0.23 & 0.32 & 289 \\
Positive & 0.38 & 0.40 & 0.39 & 204 \\
\bottomrule
\end{tabular}
\caption{CNN test-set classification report ($n=1{,}275$).}
\label{tab:cnn-perclass}
\end{table}

\begin{figure}[htbp]
\centering
\includegraphics[width=\linewidth]{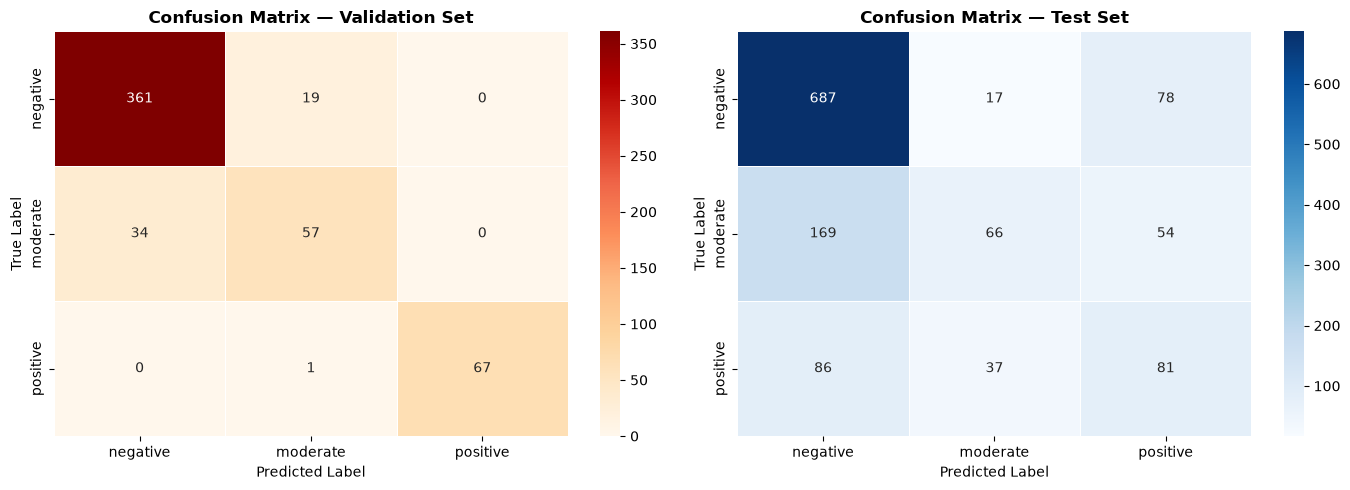}
\caption{CNN confusion matrices on the validation set (left, $n=539$) and the held-out test set (right, $n=1{,}275$).}
\label{fig:cnn-cm}
\end{figure}

\begin{figure}[htbp]
\centering
\includegraphics[width=0.75\linewidth]{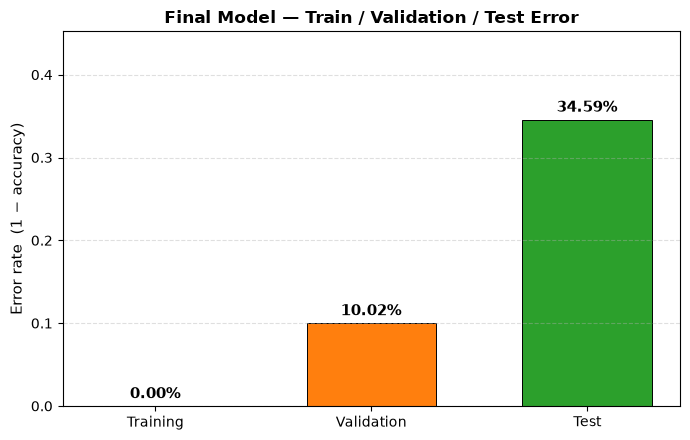}
\caption{CNN final train, validation, and test error rates, measured with a deterministic transform on all three splits.}
\label{fig:cnn-error}
\end{figure}

\begin{figure}[htbp]
\centering
\includegraphics[width=\linewidth]{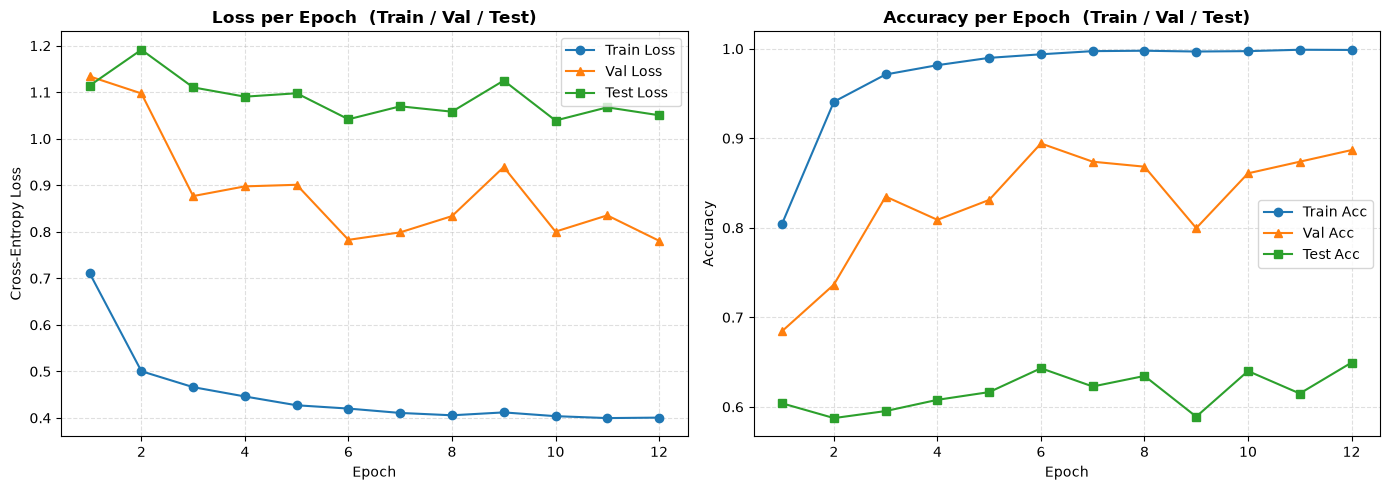}
\caption{CNN loss and accuracy per epoch (train / validation / test) during the fine-tuning stage.}
\label{fig:cnn-curves}
\end{figure}

\begin{figure}[htbp]
\centering
\includegraphics[width=0.75\linewidth]{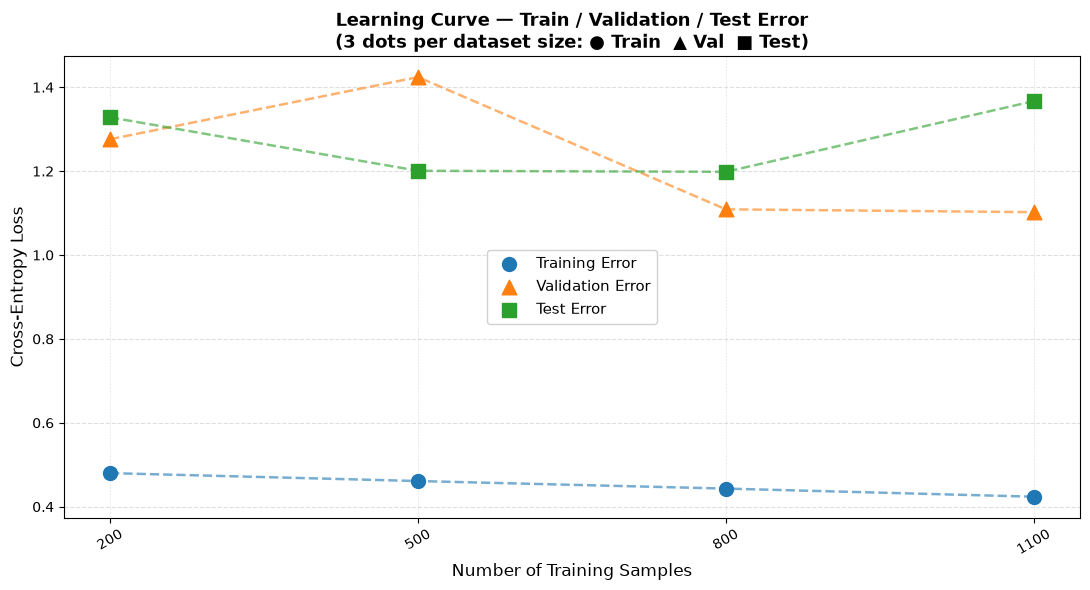}
\caption{CNN learning curve: cross-entropy loss for training, validation, and test sets as training set size grows from 200 to 1,100 samples.}
\label{fig:cnn-lc}
\end{figure}

\begin{figure}[htbp]
\centering
\includegraphics[width=\linewidth]{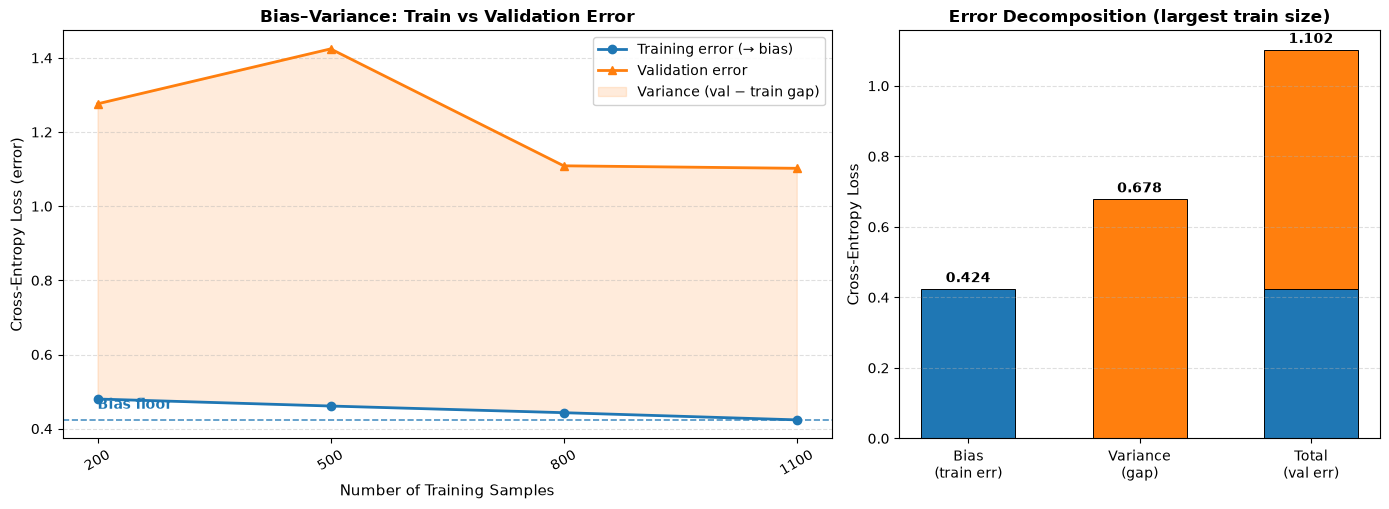}
\caption{CNN bias-variance decomposition derived from the learning-curve sweep: training error versus validation error (left), and the resulting bias/variance/total error breakdown at the largest training size tested (right).}
\label{fig:cnn-bv}
\end{figure}

Figure~\ref{fig:cnn-lc} shows the CNN's learning curve as training-set size grows, and Figure~\ref{fig:cnn-bv} decomposes the resulting error into bias and variance at the largest training size tested. The training-error floor stays low and roughly flat as more data is added, while validation and test error remain high and do not converge toward it, indicating that the model's error at this data scale is dominated by variance (overfitting to the training subjects available) rather than by insufficient model capacity.

\section{Experimental Results}
\label{sec:results}

Table~\ref{tab:comparison} summarizes all four models. The CNN achieves the highest accuracy and macro-F1 on the shared test set, followed by the Random Forest ensemble and the balanced SVM ensemble, which perform similarly to one another; the raw-pixel MLP, evaluated on its own much smaller test set, trails all three. Because the MLP's evaluation set differs in both size and composition from the other three, its accuracy figure should be read as indicative of relative model strength rather than as a directly comparable number.

\begin{table}[htbp]
\centering
\small
\begin{tabularx}{\linewidth}{X c c c c c c}
\toprule
\textbf{Model} & \textbf{Test $n$} & \textbf{Train Acc} & \textbf{Val Acc} & \textbf{Test Acc} & \textbf{Macro-F1} & \textbf{Wtd-F1} \\
\midrule
MLP (raw pixels) & 56 & 89.3\%$^{*}$ & 42.1\% & 50.0\% & 0.42 & 0.52 \\
Random Forest ensemble & 1,275 & 100.0\% & 62.2\% & 59.9\% & 0.44 & 0.57 \\
Balanced SVM ensemble & 1,275 & 100.0\% & 64.2\% & 58.2\% & 0.45 & 0.57 \\
CNN (ResNet18, fine-tuned) & 1,275 & 100.0\% & 90.0\% & \textbf{65.4\%} & \textbf{0.50} & \textbf{0.62} \\
\bottomrule
\end{tabularx}
\caption{Comparison of all four models. The Random Forest ensemble, the balanced SVM ensemble, and the CNN share the identical 1,275-image test set; the MLP uses its own, independently constructed 56-image real-only test set. $^{*}$The MLP training accuracy is measured at the final fine-tuning epoch under live augmentation rather than through a deterministic re-evaluation, and is therefore not directly comparable to the other three rows.}
\label{tab:comparison}
\end{table}

Table~\ref{tab:allperclass} restates the per-class precision, recall, and F1-score for all four models together, drawn from Tables~\ref{tab:mlp-perclass}, \ref{tab:rf-perclass}, \ref{tab:svm-perclass}, and \ref{tab:cnn-perclass}. Every model recovers the negative class most reliably, with recall ranging from 0.54 (MLP) to 0.88 (CNN). The positive (violation) class is handled inconsistently across models, with recall ranging from 0.20 (Random Forest) to 0.55 in an earlier evaluation of the same architecture (not shown here), while in the present run the CNN reaches 0.40 recall and the balanced SVM reaches 0.32. The moderate class is the hardest for every model, with F1-scores between 0.24 and 0.39, reflecting both its inherent visual ambiguity by construction and its smaller share of the labeled data relative to the negative class.

\begin{table}[htbp]
\centering
\small
\begin{tabular}{llcccc}
\toprule
\textbf{Model} & \textbf{Class} & \textbf{Precision} & \textbf{Recall} & \textbf{F1} & \textbf{Support} \\
\midrule
\multirow{3}{*}{MLP ($n=56$)}
 & Negative & 0.76 & 0.54 & 0.63 & 35 \\
 & Moderate & 0.29 & 0.58 & 0.39 & 12 \\
 & Positive & 0.29 & 0.22 & 0.25 & 9 \\
\midrule
\multirow{3}{*}{Random Forest ($n=1{,}275$)}
 & Negative & 0.69 & 0.82 & 0.75 & 782 \\
 & Moderate & 0.42 & 0.28 & 0.34 & 289 \\
 & Positive & 0.27 & 0.20 & 0.23 & 204 \\
\midrule
\multirow{3}{*}{Balanced SVM ($n=1{,}275$)}
 & Negative & 0.70 & 0.79 & 0.74 & 782 \\
 & Moderate & 0.27 & 0.22 & 0.24 & 289 \\
 & Positive & 0.40 & 0.32 & 0.36 & 204 \\
\midrule
\multirow{3}{*}{CNN ($n=1{,}275$)}
 & Negative & 0.73 & 0.88 & 0.80 & 782 \\
 & Moderate & 0.55 & 0.23 & 0.32 & 289 \\
 & Positive & 0.38 & 0.40 & 0.39 & 204 \\
\bottomrule
\end{tabular}
\caption{Per-class precision, recall, and F1-score on each model's held-out test set.}
\label{tab:allperclass}
\end{table}

\begin{figure}[htbp]
\centering
\includegraphics[width=\linewidth]{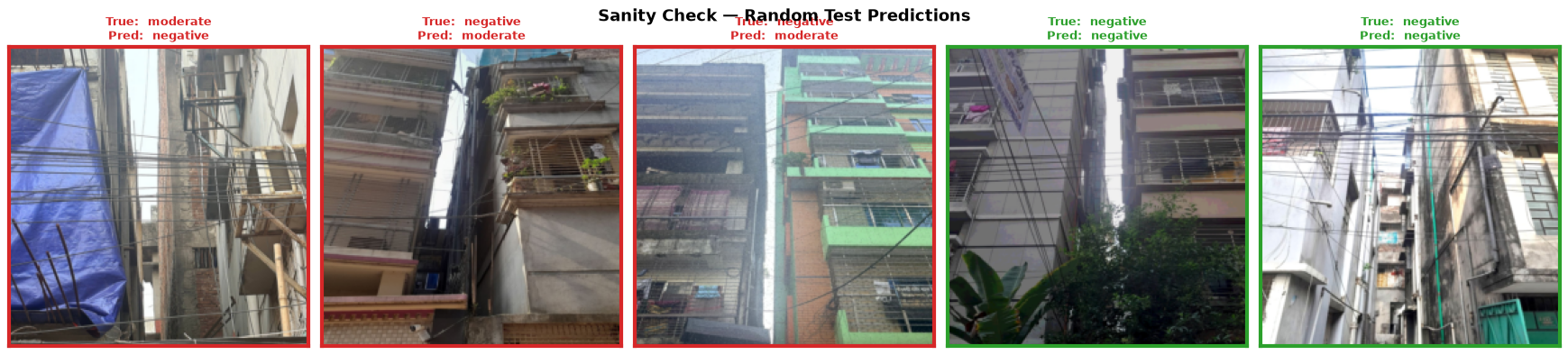}
\caption{CNN sanity check: predictions on randomly sampled test images.}
\label{fig:cnn-sanity}
\end{figure}

\begin{figure}[htbp]
\centering
\includegraphics[width=\linewidth]{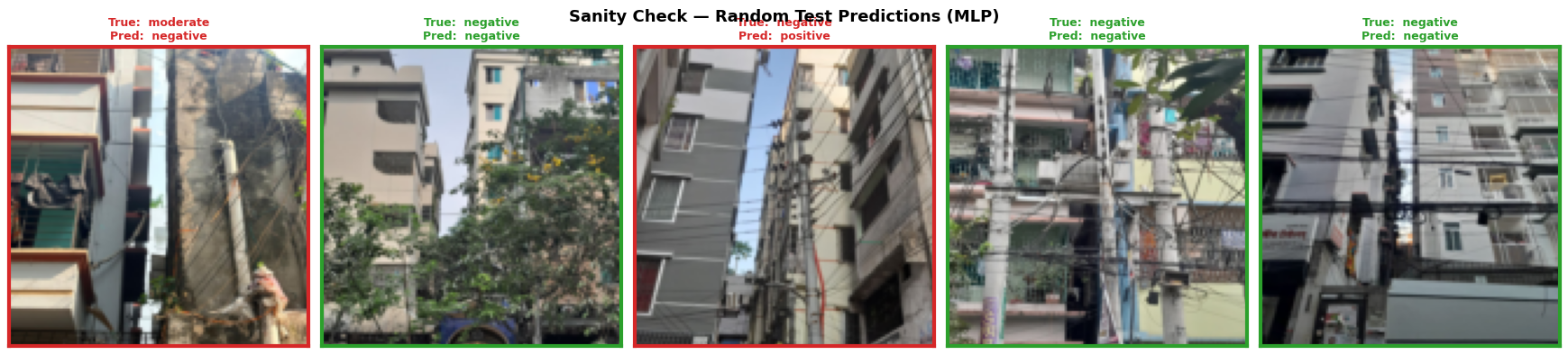}
\caption{MLP sanity check: predictions on randomly sampled test images.}
\label{fig:mlp-sanity}
\end{figure}

We also compare each model's diagnostic sanity check, a small set of test predictions with true and predicted labels overlaid (Figures~\ref{fig:cnn-sanity} and \ref{fig:mlp-sanity}). In both cases, the negative class is predicted correctly for most sampled images, and the visible errors mostly involve the moderate class being mistaken for negative, consistent with the confusion matrices reported above.

\section{Discussion}

The four models trace a clear progression in how an image is represented before classification: raw pixels, frozen embeddings passed to a classical ensemble, and features fine-tuned end to end. Test accuracy rises along this progression, from 50.0\% for the raw-pixel MLP, to 58--60\% for the two classical ensembles built on frozen embeddings, to 65.4\% for the end-to-end fine-tuned CNN. This ordering suggests that most of the benefit in this task comes from having spatial, convolutional structure in the representation at all; a further, smaller benefit comes from allowing part of the network to adapt to the target domain rather than keeping the entire backbone frozen.

The two classical ensembles, despite using different feature extractors, different base learners, and different imbalance-handling strategies, reach test accuracy within 1.7 percentage points of each other (59.9\% for Random Forest, 58.2\% for the balanced SVM), and their macro-F1 scores (0.44 and 0.45) are close as well. This suggests that, once the representation is fixed to a frozen convolutional embedding, the choice between a tree-based ensemble and a linear-kernel bagged SVM makes comparatively little difference to overall performance. The two ensembles do differ in one useful respect: the SVM's exact class-balancing strategy trades some flexibility for a lower-variance validation bias-variance profile (Figure~\ref{fig:svm-bv}) compared with the Random Forest ensemble's higher-variance profile (Figure~\ref{fig:rf-bv}), even though neither improves test accuracy over the other.

Every model, regardless of representation, struggles most with the moderate class. This is expected, since the class was defined, by construction, to capture cases where even a human annotator could not confidently assign negative or positive; asking a classifier to draw a sharp decision boundary through what is explicitly an ambiguity region is a difficult objective. The moderate class is also the smallest of the three (82 of 372 labeled images, 22\%), which compounds the difficulty. Both factors point toward the same practical remedy: collecting more labeled real images in the moderate range, and considering whether some borderline cases might be better served by a continuous compliance score, derived from the measured setback distance, rather than a hard three-way label.

The CNN's error profile also deserves comment. Its training error is exactly zero while its test error is 34.6\%, and its learning curve (Figure~\ref{fig:cnn-lc}) shows validation and test error remaining high and non-decreasing as more training data is added, while training error stays low and flat. Both observations point to variance, rather than insufficient model capacity, as the dominant source of the CNN's remaining error, a pattern also visible, to a lesser degree, in the MLP's own loss curves (Figure~\ref{fig:mlp-curves}). Given this, further gains for the CNN are more likely to come from stronger regularization, more real training images, or a smaller number of fine-tuned parameters than from a larger backbone.

A limitation shared by all four models is that the test sets, while leakage-free at the subject level, still contain a large share of synthetically augmented images (1,071 of the 1,275 test images used for the CNN, Random Forest, and SVM models); accuracy on real, unaugmented photographs alone is likely somewhat different from the headline figures reported here, since augmented copies of the same building, even when confined entirely to one split, are visually more similar to one another than independent real photographs would be. A second limitation is that the MLP's much smaller, real-image-only test set (56 images) makes its accuracy estimate considerably noisier than the 1,275-image test set shared by the other three models, so the precise 50.0\% figure should not be over-interpreted at the level of individual percentage points.

\section{Conclusion}

This paper compares four machine learning approaches, spanning raw-pixel classification, two classical ensembles on frozen convolutional embeddings, and an end-to-end fine-tuned convolutional network, for detecting building setback violations in Dhaka from street-level photographs. Using the dataset of 372 labeled images across sixteen neighborhoods, augmented to 6,340 training-eligible images, and a leakage-free, subject-level evaluation protocol, we found that the fine-tuned ResNet18 CNN reaches the highest test accuracy (65.4\%, macro-F1 0.50) on a shared 1,275-image held-out test set, ahead of a heterogeneous Random Forest ensemble (59.9\%, macro-F1 0.44) and a class-balanced bagged SVM ensemble (58.2\%, macro-F1 0.45); a from-scratch multi-layer perceptron operating on raw pixels reached 50.0\% (macro-F1 0.42) on its own smaller test set. All four models classify compliant buildings reliably but perform noticeably worse on the moderate, borderline class, which is both the smallest class in the dataset and, by its definition, the hardest to separate from the other two even for a human annotator.

These results indicate that convolutional feature representations, whether frozen or fine-tuned, are necessary for this task, and that fine-tuning provides a further, more modest improvement over frozen embeddings. They also suggest that the classifier placed on top of a fixed representation matters comparatively little: a Random Forest ensemble and a bagged linear SVM, built on different embeddings with different imbalance-handling strategies, reach similar accuracy. Beyond the specific numbers, the dataset, annotation protocol, and leakage-free evaluation methodology established here provide a reproducible basis for future work, including deeper convolutional backbones, semi-supervised use of the many unlabeled street images available in Dhaka, and a continuous, distance-based compliance score as an alternative to the three-way label used in this study. Because the underlying data collection method, ordinary smartphone photographs taken during field surveys, requires no specialized equipment, the approach could be integrated into RAJUK's existing inspection workflow as a screening tool that helps field staff prioritize which buildings to visit in person.

\section*{Code and Data Availability}

\subsection{Code}The source code for all four models, including data preparation, training, and evaluation scripts, is privately
 available at \textcolor{blue}{\url{https://github.com/TAUSEEF-01/Machine-Learning-Lab-Project/tree/main/final_files}} and the code may be made available on request.

\subsection{Dataset}

The image dataset (raw and augmented images) is available at
\textcolor{blue}{\url{https://doi.org/10.17632/s5fwtrvfhx.1}}

\vspace{2em}
 
\noindent\textbf{AI Usage Declaration.}
During the preparation of this manuscript, AI tools were used to support the drafting, language refinement, proofreading, and document formatting, including the organization of tables and figure captions. These tools were used solely to improve clarity, grammar, wording, and overall presentation.

\end{document}